%% file: example_paper.tex
\documentclass{article}
\usepackage[accepted]{icml2020}
\usepackage{epsfig}
\usepackage{graphicx}
\usepackage{wrapfig}
\usepackage{subfig}
\usepackage[belowskip=0pt,aboveskip=0pt,font=small]{caption}
\usepackage{microtype}
\usepackage{graphicx}
\usepackage{booktabs} 

\usepackage{amsmath}
\usepackage{amsfonts}
\usepackage{amssymb}
\usepackage{hyperref}

\input{definitions.tex}

\newcommand{\qb}{\mathbf{q}}
\newcommand{\pb}{\mathbf{p}}

\newcommand{\kl}{\mathbb{D}_{KL}}
\newcommand{\teacher}{\mathbf{q}_{\text{teacher}}}
\newcommand{\learner}{\mathbf{p}_{\text{learner}}}
\newcommand{\qteacher}{\qb_{\phi}(\yb)}
\newcommand{\plearner}{\pb_{\theta}(\yb)}

\begin{document}

\twocolumn[
\icmltitle{Learn to Talk via Proactive Knowledge Transfer}



\icmlsetsymbol{equal}{*}

\begin{icmlauthorlist}
\icmlauthor{Qing Sun}{fb}
\icmlauthor{James Cross}{fb}
\end{icmlauthorlist}

\icmlaffiliation{fb}{Facebook AI}

\icmlcorrespondingauthor{Qing Sun}{sunqingtju@gmail.com}

\icmlkeywords{Machine Learning, ICML}

\vskip 0.3in
]



\printAffiliationsAndNotice{}  

\begin{abstract}
Knowledge Transfer has been applied in solving a wide variety of problems. For example, knowledge can be transferred between tasks (e.g., learning to handle novel situations by leveraging prior knowledge) or between agents (e.g., learning from others without direct experience). Without loss of generality, we relate knowledge transfer to KL-divergence minimization, i.e., matching the (belief) distributions of learners and teachers. The equivalence gives us a new perspective in understanding variants of the KL-divergence by looking at \emph{how learners structure their interaction with teachers in order to acquire knowledge}.

In this paper, we provide an in-depth analysis of KL-divergence minimization in \texttt{Forward} and \texttt{Backward} orders, which shows that learners are reinforced via on-policy learning in \texttt{Backward}. In contrast, learners are supervised in \texttt{Forward}. Moreover, our analysis is gradient-based, so it can be generalized to arbitrary tasks and help to decide which order to minimize given the property of the task. By replacing \texttt{Forward} with \texttt{Backward} in Knowledge Distillation, we observed +0.7-1.1 BLEU gains on the WMT'17 De-En and IWSLT'15 Th-En machine translation tasks.
\end{abstract}


\section{Introduction}
Knowledge transfer is a rapidly growing and advancing research area in AI. As humans, we live in a world that is 
\begin{itemize}
    \item Stochastic. We have to handle novel situations by leveraging \textit{prior knowledge}. For example, once you've learned to walk, being able to run is not far off.
    \item Partially observed. To learn efficiently, we have to \textit{share knowledge}. For example, we can learn the eating habit of cheetahs from Wikipedia rather than observing their behavior over years in Africa.
\end{itemize}
In practice, we have observed models become bigger as the performance grows.  
There are many reasons reducing the size of these models is important, such as real-time inference efficiency and deployment on mobile devices.
Transferring knowledge from large, powerful models to compact models has been proven to be a practical solution. Clearly, knowledge transfer between agents or tasks is an important component towards AGI. 
\begin{figure*}[t]
\centering
\subfloat[Knowledge Distillation (KD)]{\includegraphics[trim=0pt 0pt  0pt 0pt, clip=true, width=0.45\textwidth]{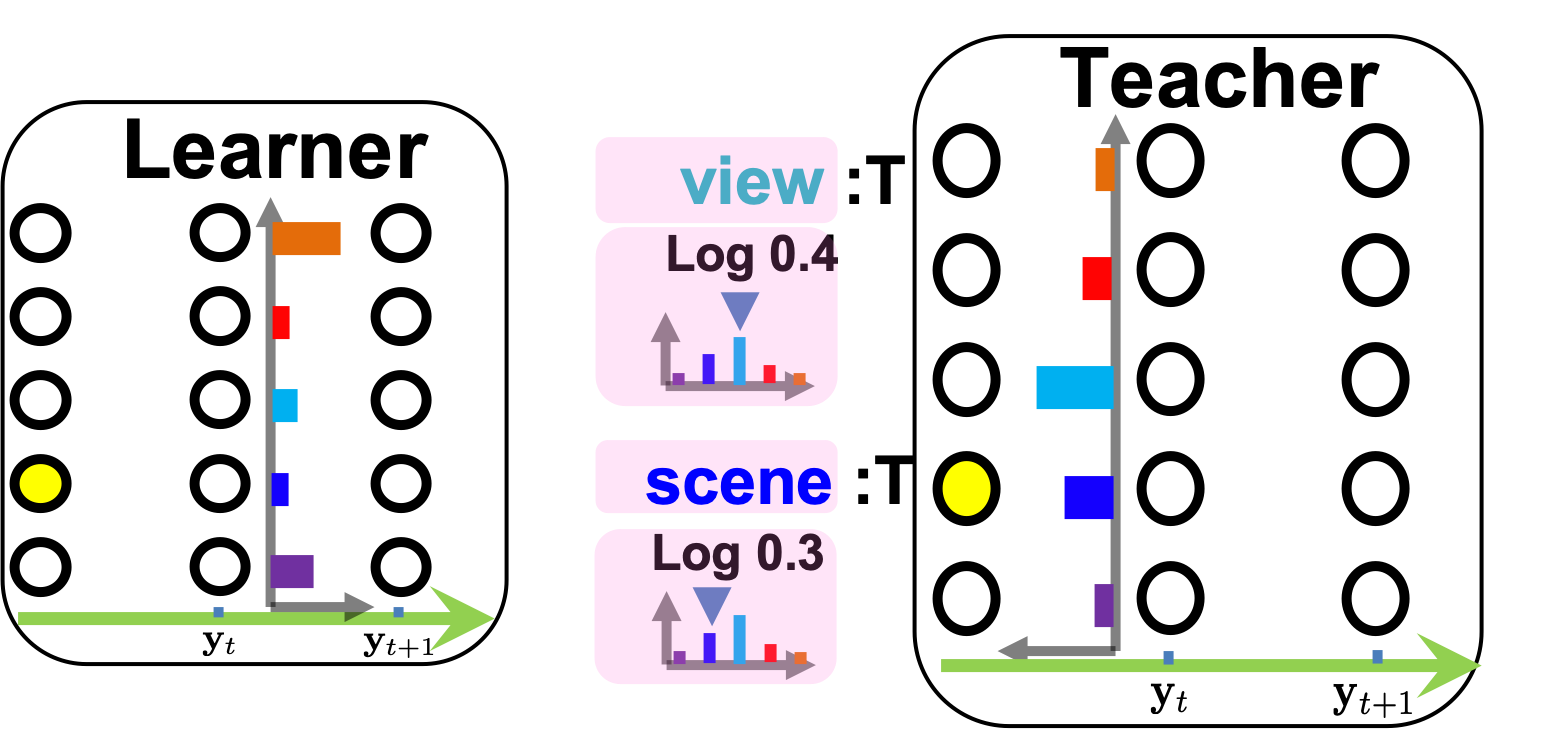}} \quad
\subfloat[Knowledge Acquisition (KA)]{\includegraphics[trim=0pt 0pt  0pt 0pt, clip=true, width=0.45\textwidth]{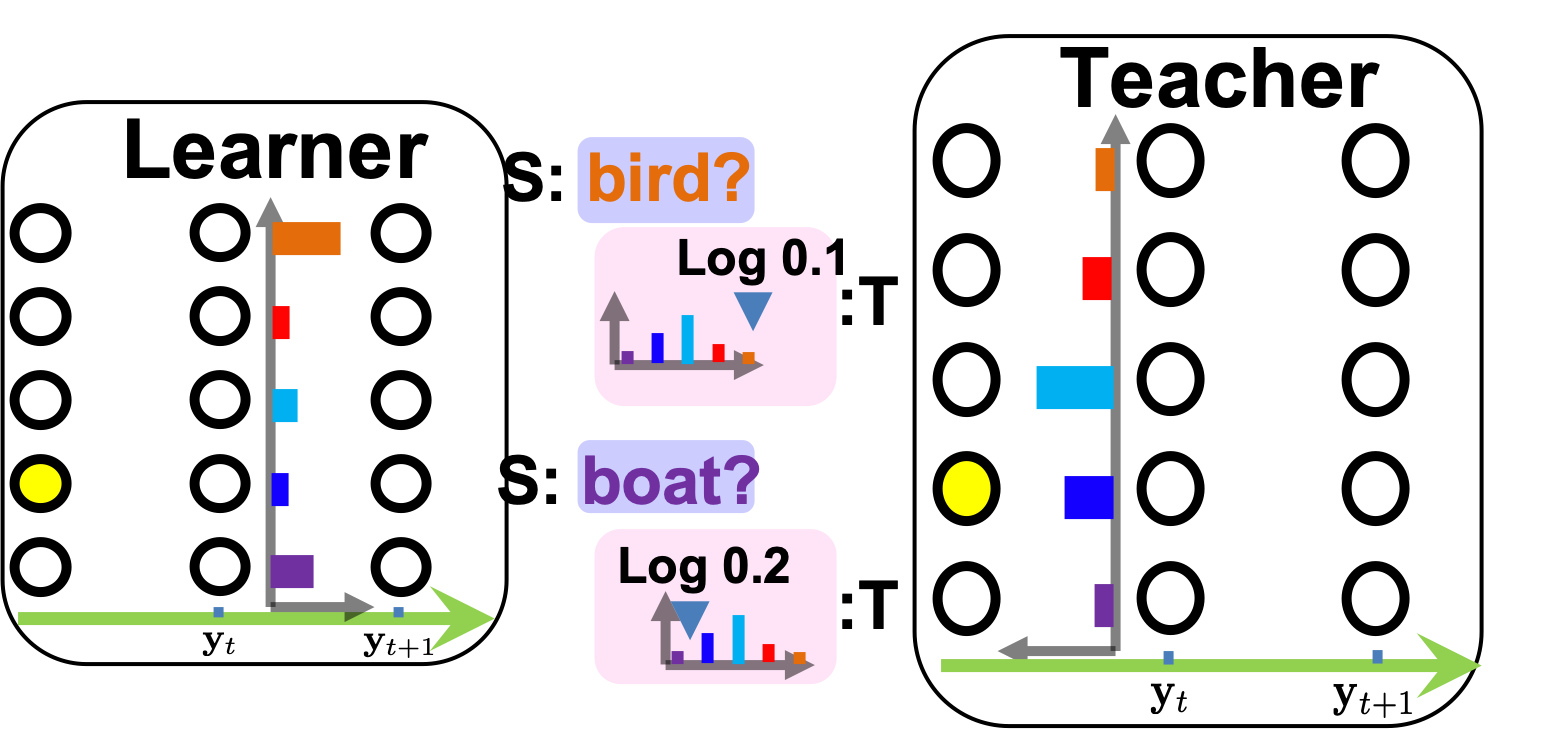}}
\caption{Learning as a dialog. (a) in KD, the teacher tells the learner to produce ``view'' since the teacher believes it has the highest probability $\qb(\text{``view''} \mid \hb_t)=0.4$, or the second best ``scene''. (b) in KA, the learner first propose to say ``bird'' because the learner thinks it's the best, but the teacher replies ``This token is bad, it's only $10\%$ likely''. Then, the learner suggests ``boat'', the teacher says ``well, it sounds better ($20\%$), but not good enough''.}
\label{fig:dialog}
\end{figure*}

What is knowledge transfer? In general, it can be regarded as a learning process by which learners acquire new or modify existing knowledge by interacting with teachers. If we represent \textit{knowledge} as a probabilistic distribution, e.g.,
\begin{align}
    \pb \big(\yb_{rain} \mid \yb_{g}=\text{``wet''} \big) =   \begin{cases}
    0.8 & \, \text{if } \yb_{train} \text{ is True}\\
    0.2 & \, \text{otherwise}
  \end{cases} \nonumber
\end{align}

Thus, learning means to get $\learner$ closer to $\teacher$, where $\learner$ and $\teacher$ are distributions of learner and teacher.

It's natural to relate knowledge transfer to KL-divergence which measures how much information (or knowledge) is lost when approximating a distribution with another. Existing approaches can be categorized as minimizing KL-divergence in \texttt{Forward} or \texttt{Backward} order. 
\begin{align}
    (a) \underbrace{\kl (\teacher \| \learner)}_{\texttt{Forward}} \text{ and }
    (b) \underbrace{\kl (\learner \| \teacher)}_{\texttt{Backward}} \nonumber
\end{align}
Learners with infinite capacity would end up behaving in the exact same way to teachers. In this case, the order doesn't matter. Unfortunately, this is not true. In practice, learners are typically restricted in a simple function family, e.g., Gaussian, or parameterized by neural networks with finite layers and hidden units. Meanwhile, teachers are not perfect. $\teacher$ is noisy especially in high-dimensional action space.

The question is which order performs better? Previous works \cite{murphy2012machine} attempt to analyze their difference by restricting $\learner$ and $\teacher$ to be Gaussian distributions. However, the learning process, i.e., how the learner interact with the teacher and acquires knowledge, is totally unclear. In this paper, we reinterpret KL-divergence minimization as knowledge transfer and provide an in-depth analysis on the way in which the learner acquires knowledge from the teacher, without applying any constraints on the distributions. 

At a high level, we noticed that 
\begin{align}
    (a) \texttt{Forward} \Rightarrow \underbrace{\text{SL}}_{\text{off-policy}} \text{ and }
    (b) \texttt{Backward} \Rightarrow \underbrace{\text{RL}}_{\text{on-policy}} \nonumber
\end{align}
\texttt{Backward} encourages learners to be exposed to their own distributions and be reinforced by on-policy learning. Thus, we propose to minimize \texttt{Backward} in solving sequential decision making problems. To verify our hypothesis, we revisit Knowledge Distillation (KD)\cite{kim2016sequence} which attempts to distill knowledge from teacher to learner by minimizing \texttt{Forward} on Neural Machine Translation (NMT) task. The goal of the task is to generate a sequence of tokens in the target language given a sentence in the source language. The generation procedure is: at position $t$, given previous actions (i.e., tokens generated so far), try to produce a token which is expected to give maximum reward (i.e., correctly translate the meaning in source sentence). Basically, NMT is to solve a sequential decision making problem with high-dimensional, discrete action space. Exposure bias is known to be an important issue due to lacking exploration. 

We simply replace \texttt{Forward} with \texttt{Backward} and call our approach Knowledge Acquisition (\textbf{KA}). We describe the learning process (supervised or reinforced) as a dialog, shown in Fig.\ref{fig:dialog}. Empirical results show +0.7-1.1 BLEU gains on WMT'17 De-En and IWSLT'15 Th-En tasks.


\section{Related Work}
Our work closely relates to three lines of work: knowledge transfer, KL-divergence and dealing with the mismatch between training and inference in sequence generation tasks. All of them have already been explored in literature.

\textbf{Transfer learning.} 
Knowledge distillation \citep{hinton2015distilling} attempts to replace hard labels with soft labels (i.e, distribution), which motivates learners to capture the relation between categories. For example, ``cat'' is more closer to ``dog'' than ``car''. L2 regularization \cite{Kirkpatrick3521} on model parameters is used to avoid catastrophic forgetting, especially when the data is not accessible in the future. In this paper, we minimize KL-divergence rather than Euclidean distance because KL originates from information theory and measures how much \textit{knowledge} is lost.

\texttt{Sequence generation.} Pre-training with monolingual data \citep{radford2018improving,radford2019language} has had significant success in language understanding. Back-translation \cite{sennrich2015improving,edunov-etal-2018-understanding} translates monolingual data in the target language into source language and uses synthetic parallel data to improve (source $\rightarrow$ target) model. Further, dual learning \cite{cheng-etal-2016-semi,hoang-etal-2018-iterative} can be thought of as bidirectional or iterative back-translation. Another line of work is knowledge distillation \cite{kim2016sequence}, where sequence-level and token-level approaches have been proposed. At sequence-level, learners are trained on the augmented dataset with outputs of running beam search with teachers. At token-level, they get the conditional probability of each token given preceding tokens closer to that of teachers. In other words, this is equivalent to minimizing KL-divergence in \texttt{Forward} order. Back-translation is a special case and performs at sequence-level, where the teacher is (target $\rightarrow$ source) model. \cite{yu2017seqgan} use adversarial training to encourage the models producing human-like sequences by learning a sequence-level discriminator to distinguish generated sequence and human references. In fact, this is equivalent to minimizing Jensen-Shannon Divergence (JSD, $0.5*$ \texttt{Forward} $+ 0.5*$ \texttt{Backward}) at sequence-level. In this paper, we provide an in-depth analysis on why \texttt{Backward} helps mitigate exposure bias. We observe that \texttt{Backward} performs the best among \texttt{Forward} and JSD.

\textbf{Order in KL-divergence.} Previous work \cite{murphy2012machine} assumes the teacher (real distribution) is multi-modal Gaussian and the learner (surrogate distribution) is uni-modal Gaussian. Minimizing \texttt{Forward} is zero-avoiding for the teacher and the resulting modes of the learner will be in low density, right between modes of the teacher [$\learner $ ``covers'' $\teacher$]. In contrast, minimizing \texttt{Backward} is zero-forcing for the learner, and the learner locks on a single mode. The insight has been widely used in a variety of research areas such as variational inference \cite{wainwright_ftml08,wainwright2008graphical} and GAN \cite{goodfellow2014generative}. However, the way in which knowledge is transferred to the learner is unclear.  Moreover, the uni-modal distribution constraint doesn’t hold anymore when the learner is parameterized by a neural network. In this paper, we relax the constraint on distributions and attempt to take a close look at the ``gradients'', which unveils the mystery in learning process and provides a guidance on which order to minimize given a specific task.

\textbf{Learning $\neq$ inference.} To handle the mismatch between training and inference, previous works attempt to directly optimize the task-specific metric at test time. \cite{ranzato2015sequence} propose sequence-level training algorithm in reinforcement learning. The models receive rewards until the completion of the entire sequence. Considering that the search space in sequence generation is exponentially large, i.e, $O(|\mathcal{V}|^T)$, where $\mathcal{V}$ is a set of tokens in the vocabulary ($\sim$ 10K or more) and $T$ is the length of the sequence ($\sim$ 20 or more), the rewards are extremely sparse. This makes the training unstable. To alleviate the sparse rewards problem, \cite{liu2017improved} use Monte Carlo rollouts to estimate immediate rewards at each position. Unfortunately, the estimation is very noisy and of high variance due to the exponentially large search space. Moreover, the training is computationally expensive. In this paper, learners are able to receive feedback at each position such that the reward is dense. In addition, teachers serve as ``Critic'' that estimate action-value function, i.e., the rewards of producing the current token given previous tokens, which only needs a single forward pass of neural network and then is computationally efficient.

To deal with exposure bias, \cite{bengio2015scheduled} propose to gradually replace tokens from human references to generated tokens during training. The training starts with tokens from human references and ends up with using generated tokens. However, the rewards which are the matching n-grams with a few human references limits the capability of the models in exploration.  In this paper, we pair the learner with a knowledgeable teacher which offers smart advice based on semantic meanings.

\section{Learning Strategy}
\subsection{Notation}
We let $\pb_{\theta}(\yb)$ denote $\learner$ and $\qb_{\phi}(\yb)$ denote $\teacher$, where $\yb=(\yb_{t}, \yb_{t+1},\dots,\yb_{t'})$ represents a sequence of variables. Without loss of generality, $\yb_t$ can be a action (e.g., ``Turn left'') or a token ( e.g., ``Amazing''). 

\subsection{Supervised versus Reinforced}
\label{sec:why}
To be simple, we define knowledge transfer as an optimization problem that attempts to estimate parameter $\theta$ by minimizing KL-divergence between $\plearner$ and $\qteacher$ with fixed $\phi$.

It is known that KL-divergence is asymmetric and the order makes a difference. Thus, we would investigate how the choice of order affects the way the knowledge is transferred.  
\begin{itemize}
    \item[(a)] Forward 
\begin{align}
   & \kl \big( \qteacher \| \plearner \big) \nonumber \\
= \, & -\mathbb{E}_{\qteacher} \big[ \log \plearner \big] - \underbrace{\mathbb{H} \big[ \qteacher \big]}_{\text{Const } w.r.t. \plearner} 
\end{align}
Empirically, we \textbf{M}aximize the \textbf{L}og-likelihood to \textbf{E}stimate parameter $\theta$
\begin{align}
\sum_{D_y \sim \qteacher}  \log \plearner  \label{eqn:forward}
\end{align}


\item[(b)]Backward 
\begin{align}
   & \kl \big( \plearner \| \qteacher) \big) \nonumber \\
= \, & -\mathbb{E}_{\underbrace{\plearner}_{\text{Policy}}} \big[ \underbrace{\log \qteacher}_{\text{Reward}: r(\textbf{y})} \big] - \mathbb{H} \big[ \plearner \big]
\end{align}
Empirically, we learn a policy that has the highest entropy while maximizing the expected reward
\begin{align}
\sum_{D_y \sim \plearner}  r(D_y) +  \mathbb{H} \big[ \plearner \big]  \label{eqn:backward}
\end{align}
\end{itemize}
From Eq.\eqref{eqn:forward} and Eq.\eqref{eqn:backward}, we can see that the learner in \texttt{Forward} is supervised while in \texttt{Backward} is reinforced.

Which order do transfer learning algorithms minimize? We can see that conventional transfer learning algorithms typically minimize \texttt{Forward}. Given the definition, some existing works can be reinterpreted as ``knowledge transfer''. For example,
\begin{itemize}
    \item Variational Inference \cite{wainwright2008graphical,kingma2013auto}. The teacher is the real posterior distribution which is complex, non-parametric and induced by Bayes' rule. The learner is a distribution either in a predefined class, e.g., Gaussian, or parameterized by a neural network. (see Appendix \ref{sec:vi})
    \item Actor-Critic \cite{konda2000actor,DBLP:journals/corr/abs-1801-01290}. With entropy constraint, the teacher is the softmax of the Q function \cite{haarnoja2017reinforcement} (or estimated rewards via Monte Carlo rollouts) and the learner is the policy. (see Appendix \ref{sec:ac})
\end{itemize}
However, they minimize \texttt{Backward}.

Then, which order to minimize? Given Eqn.\eqref{eqn:forward} and Eqn.\eqref{eqn:backward}, can we make a better decision? To answer the question, let's zoom in and see what happens during optimization.

\begin{figure*}[t]
\centering
\subfloat{\includegraphics[trim=0pt 0pt  0pt 0pt, clip=true, width=0.45\textwidth]{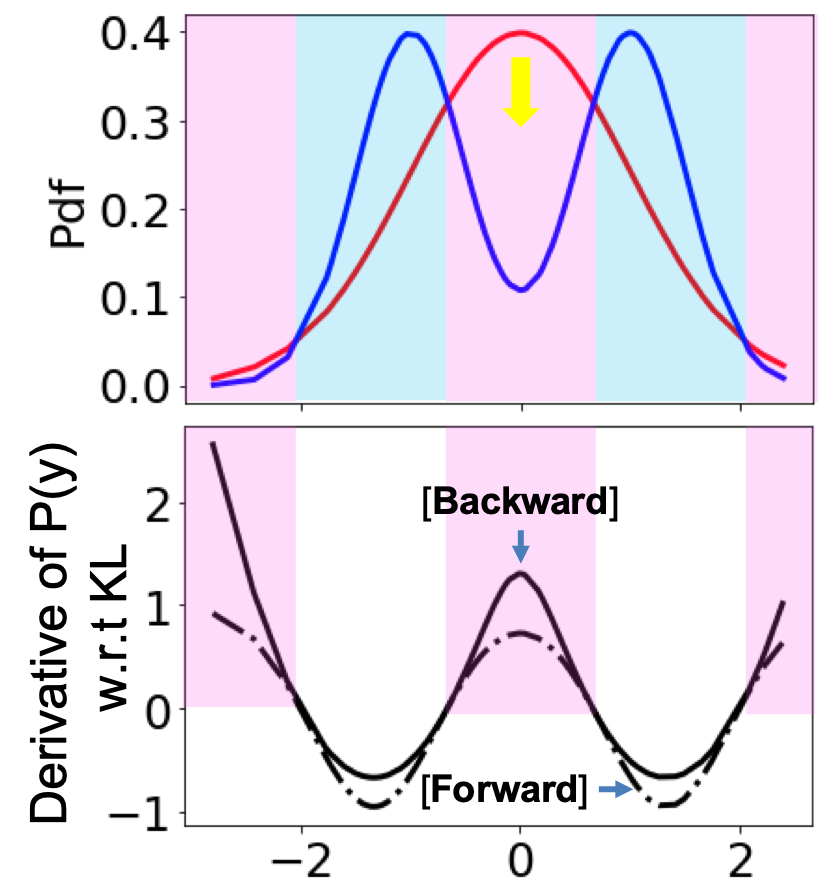}} \quad
\subfloat{\includegraphics[trim=0pt 0pt  0pt 0pt, clip=true, width=0.4\textwidth]{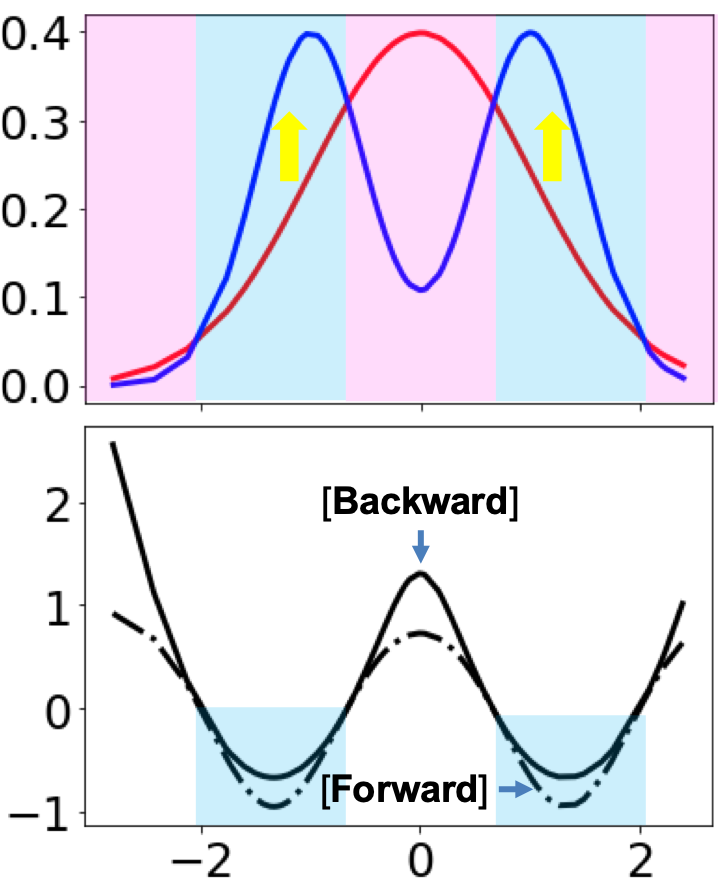}}
\caption{Derivative of $\pb(y)$ w.r.t. $\kl$ in forward and backward orders. Top row: we assume the teacher is already powerful enough to capture the complex distribution and let it be a mixture Gaussian (shown in blue). In contrast, we represent the learner as a single Gaussian (shown in red).  We partition the space $y$ into two disjoint regions: $\pb(y) > \qb(y)$ in pink (I) and $\pb(y) < \qb(y)$ in blue (II). Bottom row: let the dashed curve denote the derivative of $\kl$ in forward order and the solid curve denote the derivative of $\kl$ in backward order. We can see that in the pink region (I) on the left, \texttt{Backward} pushes down harder than \texttt{Forward}, while in the blue region (II) on the right, \texttt{Forward} pulls up harder than \texttt{Backward}.}
\label{fig:derivative}
\end{figure*}


\subsection{Derivatives}
\label{sec:dev}
Without loss of generality, we simplify the problem to a 1-D problem. Based on Lagrangian relaxation (see Appendix \ref{sec:derivative}), the derivatives of $\kl$ w.r.t. $\pb(y)$ are  

\begin{align}
    (a) \text{ Forward} \quad \quad \frac{\partial \kl(\qb(y) || \pb(y))}{\partial \mathbf{p(y)}} = \underbrace{1- \frac{\mathbf{q}(y)}{\mathbf{p}(y)}}_{G_{\qb \| \pb}(y)}
    \label{eqn:forward_deriv}
\end{align}

\begin{align}
   (b) \text{ Backward} \quad \quad \frac{\partial \kl(\mathbf{p}(y) || \mathbf{q}(y))}{\partial \mathbf{p(y)}} = \underbrace{\log \frac{\mathbf{p}(y)}{\mathbf{q}(y)}}_{G_{\pb \| \qb}(y)} \label{eqn:backward_deriv}
\end{align}
Eqn.\eqref{eqn:forward_deriv} and Eqn.\eqref{eqn:backward_deriv} tell us that
\begin{align}
    \text{(I) } & \qb(y) > \pb(x) \Rightarrow G_{\qb \| \pb}(y) <0 \text{ and } G_{\pb \| \qb}(y) < 0 \nonumber \\
    \text{(II) } & \qb(y) < \pb(x) \Rightarrow G_{\qb \| \pb}(y) >0 \text{ and } G_{\pb \| \qb}(y) > 0 \nonumber 
\end{align}

\texttt{Forward} and \texttt{Backward} always update $\pb(y)$ in the same direction. This is intuitive because they share the same goal: get $\pb(y)$ closer to $\qb(y)$. Thus, the learner has to pull up the probability of $y$ which is under-estimated, i.e., $\pb(y) < \qb(y)$, while pushing down the probability of $y$ which is over-estimated, i.e., $\pb(y) > \qb(y)$. However,
\begin{align}
  \text{In (I), } \quad \big|G_{\qb \| \pb}(y)\big| > \big| G_{\pb \| \qb}(y) \big| \nonumber \\
  \text{in (II), } \quad \big|G_{\qb \| \pb}(y)\big| < \big|G_{\pb \| \qb}(y) \big|  \nonumber
\end{align}
This means that they exert force in the same directions, but in different magnitudes (see Appendix \ref{sec:property}). In other words, \texttt{Forward} pulls up harder when the teacher thinks $y$ is good but the learner hasn't yet (in (I)). In contrast, \texttt{Backward} pushes down harder when the learner thinks $y$ is good, but teacher doesn't agree (in (II)). 

Fig.~\ref{fig:derivative} illustrates the difference in learning strategy between \texttt{Forward} and \texttt{Backward} by assuming the learner is a single-modal Gaussian and the teacher is a mixture of two Gaussians.

What does the difference mean? \texttt{Backward} encourages learners being reinforced via on-policy learning. In other words, learners explore their own distributions by trying what they believe and adjusting accordingly (in I, pink region). In contrast, \texttt{Forward} asks learners to follow and behave in a way that teachers do (in (II), blue region).

Now, let's answer the question in Sec.\ref{sec:why}. We strongly recommend \texttt{Backward} in solving sequential decision making problems where exploration is necessary, e.g., in high-dimensional action space. 

\begin{figure*}[t]
\centering
\subfloat[Forward]{\includegraphics[trim=0pt 0pt  0pt 0pt, clip=true, width=0.45\textwidth]{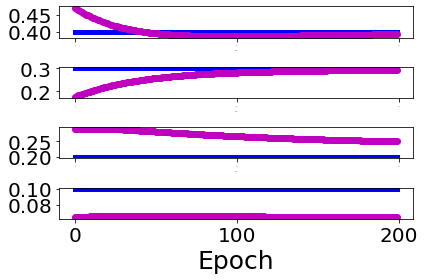}}
\subfloat[Backward]{\includegraphics[trim=0pt 0pt  0pt 0pt, clip=true, width=0.45\textwidth]{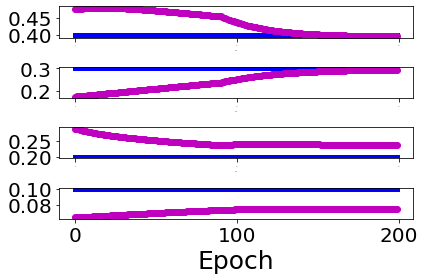}}
\caption{The evolution of $\pb(x)$ during learning. This toy task contain four discrete actions. Purple lines denote $\pb(x)$ and blue lines denote ground-truth $\qb(x)$ (aka, teacher). $(\pb_0, \qb_0) \sim (\pb_3, \qb_3)$ are shown from top to bottom (Top row: $(\pb_0, \qb_0)$). $\pb_0$ and $\pb_1$ in \texttt{Forward} converge to $\qb_0$ and $\qb_1$ in $\sim$ 30 epochs, while $\pb_2$ and $\pb_3$ stay far away from the ground-truth $\qb_2$ and $\qb_3$ with respect to those in \texttt{Backward}. }
\label{fig:toy}
\end{figure*}

To illustrate our intuition, we design a very simple task. Instead of using high-dimensional action space, we restrict the dimension to be 4, but the state is partially observed, i.e., only top-$k$ to be available, such that completing the task requires exploration. The $\kl$ term becomes
\begin{align}
    \sum_{y \in \text{Top-}k \,\, \pb(y)} \pb \big(y) \big(\log \pb \big(y \big) - \log  \qb \big(y \big) \big) \nonumber
\end{align}
where, $\text{Top-}k \, \pb (y)$ is a set of $k$ items with the highest ranking. Only $\pb(y)$ and  $\qb(y)$ according to the $y$ in the top-$k$ are calculated while the rest is discarded. Assume $y$ is a discrete variable and takes values 0, 1, 2, and 3. We build a quite simple model with a single softmax layer to produce a distribution over $y$, and set top-$k$=2. The real distribution is
\begin{align}
    \qb_0=0.4, \, \qb_1=0.3, \, \qb_2=0.2, \, \qb_3=0.1 \nonumber
\end{align}

Fig.\ref{fig:toy} shows that \texttt{forward} drives $\pb_0$ and $\pb_1$ to the real values very quickly, but almost ignore $\pb_2$ and $\pb_3$ since they are not directly optimized. In contrast, \texttt{backward} first optimizes $\pb_0$ and $\pb_2$ which are the top-2. After $\sim 100$ epochs, when $\pb_0$ and $\pb_1$ become the top-2, the loss drops very fast and even lower than that of \texttt{forward}. Moreover, \texttt{backward} drives $\pb_2$ and $\pb_3$ much closer to the real values since the model has explored 3 states, i.e., $y=0, 1, 2$.


\section{Sequence Generation as Decision Making}
Let $(\xb, \yb) \in (\mathcal{X}, \mathcal{Y})$ be an example translation pair, where $\xb$ is the source sentence and $\yb$ is the target sentence. $\yb_{t}$ is the token at position $t$ and $\yb_{<t}$ is tokens before position $t$. \cite{sutskever2014sequence} propose sequence-to-sequence models which typically factorize the joint probability $\pb(\yb \mid \xb)$ over a sequence of conditional probabilities with parameter $\theta$:
\begin{align}
\pb \big(\yb | \xb \big) = \prod_{t=0}^T \pb_{\theta} \big(\yb_t \mid \yb_{<t}, \xb \big)
\label{eqn:seq2seq}
\end{align}

where, $\pb_{\theta}(\yb_t \mid \yb_{<t}, \xb)$ is the probability of the token $\yb_t$ given previous tokens $\yb_{<t}$ and the source sentence $\xb$. Basically, the preceding tokens $
\yb_{<t}$ are encoded into the hidden states via a state transition function 
\begin{align}
\hb_t = f(\hb_{t-1}, \yb_{t-1}; \xb)
\label{eqn:world}
\end{align}
By substituting Eqn.\eqref{eqn:world} for Eqn.\eqref{eqn:seq2seq}, we have
\begin{align}
\pb \big(\yb | \xb \big) = \prod_{t=0}^T \underbrace{\pb_{\theta}(\yb_t \mid \hb_t)}_{\text{Policy } \pi}
\end{align}
This tells us that the sequence models, e.g., RNNs, acts like a stochastic policy which picks a discrete action, i.e., producing a token $\yb_t \in \mathcal{V}$, running on a world model $\mathcal{M}$ with transition function $f$. 

\textbf{Training.} We minimize the cross-entropy loss
  \begin{align}
      \mathcal{L}_{CE}= -\sum_t \log \pb_{\theta} \big(\yb_t^{ref} \mid \yb_{<t}^{ref}, \xb \big)
  \end{align}
where, $\yb^{ref}$ denotes human references. At each position, the models are conditioned on the ground-truth tokens annotated by humans no matter what tokens are predicted by themselves.

\textbf{Inference.} In sequence generation tasks, exact inference is intractable due to exponentially large search space. Instead, we apply an approximation inference algorithm - Beam Search (BS). BS is a greedy heuristic search that maintains the top-B most likely partial sequences through the search tree, where B is referred to as the beam size. At each position, BS expands these B partial sequences to all possible beam extensions and then selects the B highest scoring among the expansions. Unlike training, ground-truth tokens are not available. The models have to use their own predictions in decoding.

\textbf{Evaluation.} To evaluate the quality of generated sequences, we typically use metrics such as BLEU score to measure their n-gram overlap with human references. However, the standard metrics are problematic and none of them correlate strongly with human evaluation at the level of individual sentences. For example, given a human reference ``Amazing view along the sea'', the sequence ''The scenery of the seaside is beautiful'' gets low BLEU score because there is no matching n-grams of order 2, 3, or 4. In addition, human references are a few sentences lacking in diversity. For example, when asking more people, they might say ``A nice beach'' or `` what amazing view of the seashore''.

\subsection{Discrepancy among procedures}
\label{sec:bias}
\textbf{Exposure bias.} During training, the models only explore the training data distribution, but never get exposed to their own predictions. Searching in under-explored space causes errors. More importantly, such errors accumulate over time because of greedy search. Given a example with ground-truth `` Amazing view along the sea'', assume there are no sentences in training data starting with ``Amazing''.
\begin{align}
    \text{$t_1$: } & \text{pick a token from } \pb(\yb | \text{``Amazing''}, \xb) \Rightarrow \text{``cup''} \nonumber \\
    \text{$t_2$: } & \text{pick a token from } \pb(\yb | \text{``Amazing cup''}, \xb) \Rightarrow \text{``on''} \nonumber \\
    \text{$t_3$: } & \text{pick a token from } \pb(\yb | \text{``Amazing cup on''}, \xb) \Rightarrow \text{``table''} \nonumber
\end{align}
We can see that the poor token ``cup'' caused by the noisy distribution $\pb(\yb | \text{``Amazing''}, \xb)$ makes the situation even worse. The distributions become more and more noisy and the generation goes far away.

\textbf{Sub-optimal models.} The training objective is different from the metrics used in evaluation. To address this issue, some works attempt to directly optimize the metrics. It definitely helps improve the score, but we suspect that this might hurt the models because poor metrics discourage learning the semantic meanings. For example, the low score of ``The scenery of the seaside is beautiful'' could make the representation of ``Amazing'' far from that of ``beautiful'' or ``seaside'' far from that of ``sea''.

\subsection{Proactive Knowledge Transfer}
Given a teacher with probability $\qb_{\phi}(\yb_t \mid \hb_t)$, previous work \cite{kim2016sequence} attempts to distill knowledge by minimizing \texttt{Forward}. However, exposure bias in Sec \ref{sec:bias} motivates us to leverage on-policy learning which allows learners to be exposed to their own distributions and then reduce accumulated errors. Therefore, we propose to minimize \texttt{Backward} in sequence generation tasks. More concretely, we minimize
\begin{align}
    \sum_t \kl \Big(\pb_{\theta}( \yb_t \mid \hb_t) \| \qb_{\phi}( \yb_t \mid \hb_t) \Big)
\end{align}
i.e., learners learn to talk by actively asking teachers for advice at each position $t$. 

Our approach is actually a kind of reverse Knowledge Distillation (\textbf{KD}), which we call knowledge Acquisition (\textbf{KA}). 

We describe the learning process via a dialogue between teacher and learner, shown in Fig.\ref{fig:dialog}.

Unlike metrics, e.g., BLEU score, $\qb_{\phi}( \yb_t \mid \hb_t)$ would give higher rewards to tokens with similar semantic meaning. For example, given the ground-truth token ``scene'', ``view'' receives low BLEU score, but high probability from teachers. This would encourage learners to produce diverse translations. In addition, teachers can emit reward at each position rather than waiting for the completion of the entire sequence, which makes the optimization much stable.

\section{Experiments}

\begin{table*}[t]
  \centering 
  \small\addtolength{\tabcolsep}{1pt}
  \caption{Model configurations for WMT'17 De-En and IWSLT'15 Th-En}\label{tab:config}
  \begin{tabular}[t]{ccccccccccc} 
    \toprule 
     &  & \multicolumn{3}{c}{{Encoder}}  &  \multicolumn{3}{c}{{Decoder}} \\ [0.2ex]
    \cline{3-8} 
    &  & layer & dim & head  &  layer & dim & head \\ [0.2ex]
    \midrule
    \multicolumn{1}{c|}{{De-En}} & learner & 6 & 512 & 4  & 1 & 512 & 4 & \\
    \multicolumn{1}{c|}{} & teacher & 6 & 1024 & 16 & 6 & 1024 & 16  \\
    \multicolumn{1}{c|}{{Th-En}} & learner & 1 & 128 & 1 & 1 & 128 & 1  \\
    \multicolumn{1}{c|}{} & teacher & 3 & 256 & 2 & 1 & 256 & 2\\ [0.5ex]
    \bottomrule 
  \end{tabular} 
\end{table*}

\begin{figure*}[t]
\centering
\subfloat[Accuracy vs. Top-$k$]{\includegraphics[trim=0pt 0pt  0pt 0pt, clip=true, width=0.45\textwidth]{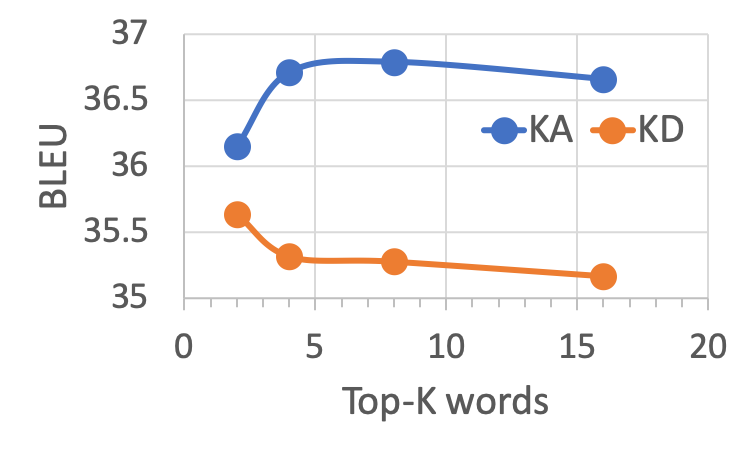} \label{fig:topk}} \quad \quad
\subfloat[No. of novel tokens]{\includegraphics[trim=0pt 0pt  0pt 0pt, clip=true, width=0.40\textwidth]{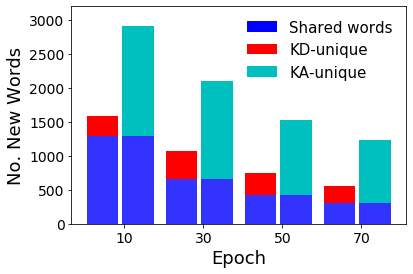} \label{fig:novel}}
\caption{Ablation study on WMT'17 De-En task. (a) Accuracy on validation set with variable $k$. Larger $k$ means a larger part of distribution is observed. (b) Number of novel tokens emerging in the top-16. High numbers indicates strong exploration in search space. We noticed that $\sim$80\% probability mass are put on the top-16 tokens.}
\end{figure*}


\begin{table*}[t]
  \centering 
  \small\addtolength{\tabcolsep}{0.5pt}
  \caption{BLEU scores on WMT'17 De-En and IWSLT'15 Th-En. (On test set)}\label{tab:bleu}
  \begin{tabular}[t]{cccccc} 
    \toprule 
     & Learner & Teacher & KD & \textbf{KA} & 1/2KA+1/2KD \\ [0.2ex]
    \midrule 
     De-En & 30.23 & 34.10 & 31.13 & \textbf{32.24} & 31.51 \\
     Th-En & 12.37 & 17.55 & 12.70 & \textbf{13.36}  & 12.93 \\ [0.5ex]
    \bottomrule 
  \end{tabular}
\end{table*}
\subsection{Dataset}
 We evaluate our approach on WMT 2017 German-English with 4M sentence pairs, validate on newstest2016 and test on newstest2017. All the
sentences are first tokenized with Moses tokenizer and then segmented into 40K joint source and target byte-pair encoding tokens \citep{sennrich2015neural}. Another Thai-English dataset comes from IWSLT 2015. There are 90k sentence pairs in training and we take 2010/2011 as the dev set and 2012/2013 as the test set. Byte-pair encoding vocabulary is of size 10K. 

\subsection{Training}
Without a good starting point, the performance of minimizing \texttt{Backward} degrades because the models are more than likely stuck on the current best tokens and unlikely to explore. We therefore pre-train learner models and fine-tune them in all the experiments by minimizing
\begin{align}
   \mathcal{L}_{ALL}(\yb; \xb, \theta, \phi) = (1-\lambda) \mathcal{L}_{NLL}(\yb;\xb, \theta) + \nonumber \\
   \lambda \sum_t \underbrace{\kl
\Big(\pb_{\theta}(\yb_t \mid \hb_{t}) \| \qb_{\phi}(\yb_t \mid \hb_{t})\Big)}_{\kl(\qb_{\phi} \| \pb_{\theta}) \text{for KD}}
\label{eqn: obj}
\end{align}

where, $\lambda$ are trade-off parameters. Basically, learners and teachers can be optimized simultaneously. However, in this paper, we simply freeze teacher models and leave joint training to future work.

\textbf{Model.} Our teacher models and learner models all use the transformer architecture, which has achieved state-of-the-art accuracy and is widely used in recent NMT research. Model configurations are listed in Table.\ref{tab:config}. We train all transformer models using the implementation in \texttt{Fairseq} \citep{ott2019fairseq}. We use Adam optimizer \citep{kingma2014adam} with $\beta_0=0.9$, $\beta_1=0.98$, $\epsilon=1e^{-8}$. Learning rate is 0.0001 and dropout rate is 0.3. At inference time, we use beam search with a beam size of 6 and length penalty 1.0.

\subsection{Results}
Our results on NMT tasks are reported in Table.\ref{tab:bleu}. We tune hyper-parameters $\lambda$ and top-$k$ on validation set where $\lambda=0.5$ for KD which is consistent with that in \cite{kim2016sequence}. We observed that KA consistently outperforms KD on both De-En (high-resource) and Th-En (low-resource) tasks by 0.7 - 1.1 BLEU score. In addition, we also test JSD (i.e., $\frac{1}{2}$ KA + $\frac{1}{2}$ KD), which is equivalent to GAN. The accuracy lies between KA and KD. The results say that KA does help to avoid exploration bias and further close the gap between training and inference.

\textbf{Exploration.} Similar to Sec.\ref{sec:dev}, we evaluate the performance by allowing only the top-$k$ tokens to be available. We vary $k$ from 2 to 16 and conduct the experiments on validation set. In Fig.\ref{fig:topk}, we see that the distribution is noisy because the accuracy of either KD or KA eventually goes down and when using the full information, i.e, $k=|\mathcal{V}|$, KD achieves 35.41 BLEU while KA achieves 35.79 BLEU, which are far away from the best. Moreover, KA is able to learn more from noise because KD directly goes down as $k$ increases, while the accuracy of KA goes up first and then drops after $k=8$.

\begin{figure}
\centering
\includegraphics[trim=0pt 0pt 0pt 0pt, clip=true, width=0.4\textwidth]{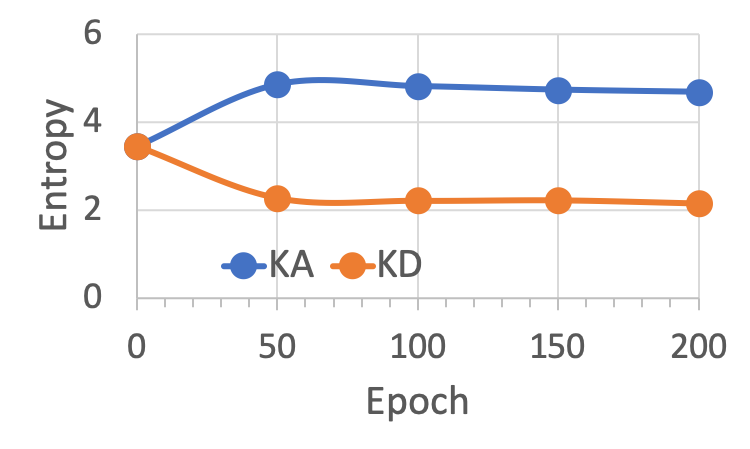}
\caption{Entropy over epochs.}
\label{fig:ent}
\end{figure}

\begin{table*}[t]
  \centering 
  \small \addtolength{\tabcolsep}{1pt}
  \caption{Top-10 tokens in KD and KA over epochs (Top $\rightarrow$ bottom). [\texttt{A}] denotes the token the models are trying to produce at the current position and \textbf{B} is the unique token included in one approach but not the other.}\label{tab:example}
  \begin{tabular}[t]{ccccccccccccccccccccccccccccccccccc} 
    \toprule 
   & \multicolumn{11}{l}{{SRC: Ich werde mich in dieser Woche darum kümmern.}} & & \multicolumn{11}{l}{{SRC:Er könnte sich nicht frei bewegen.}} \\ [0.2ex]
   & \multicolumn{11}{l}{{TRG: I will [\texttt{attend}] to it this week}} & & \multicolumn{11}{l}{{TRG: He could [\texttt{not}] move freely.}} \\ [0.2ex]
   \midrule 
   \multicolumn{1}{c|}{{KD}} & \multicolumn{11}{l}{{look, be, take, deal, do, work, care, consider, address, make}} &  & \multicolumn{11}{l}{{not, be, do, move, 't, never, hardly, avoid, have, remain}} \\
  \multicolumn{1}{c|}{} & \multicolumn{11}{l}{{look, be, take, deal, work, do, make, consider, care, get}} & & \multicolumn{11}{l}{{not, be, do, move, no, avoid, never, \textbf{stop}, have, also}} \\
  \multicolumn{1}{c|}{} & \multicolumn{11}{l}{{look, be, take, deal, work, do, care, make, consider, \textbf{concern}}} & & \multicolumn{11}{l}{{not, be, avoid, do, move, no, never, \textbf{stop}, make, have}} \\
  \multicolumn{1}{c|}{} & \multicolumn{11}{l}{{look, be, take, deal, make, work, do, care, get,\textbf{concern}}} & & \multicolumn{11}{l}{{not, be, move, do, never, no, avoid, have, make, \textbf{stop}}} \\
  \multicolumn{1}{c|}{} & \multicolumn{11}{l}{{look, be, take, deal, work, do, make, care, get, \textbf{concern}}} & & \multicolumn{11}{l}{{not, be, move, do, avoid, no, make, never, have, also}}\\
  \midrule
  \multicolumn{1}{c|}{{KA}} & \multicolumn{11}{l}{{look, be, take, deal, do, work, care, consider, address, make}} & & \multicolumn{11}{l}{{not, be, do, move, 't, never, hardly, avoid, have, remain}} \\
\multicolumn{1}{c|}{} & \multicolumn{11}{l}{{look, take, be, deal, care, work, do, make, \textbf{try}, consider}} & & \multicolumn{11}{l}{{not, be, move, avoid, never, do, no, \textbf{fail}, have, \textbf{resist}}} \\
\multicolumn{1}{c|}{} & \multicolumn{11}{l}{{take, look, be, deal, care, work, make, consider, do, \textbf{give}}} & & \multicolumn{11}{l}{{not, move, be, never, hardly, avoid, do, make, no, also}} \\
\multicolumn{1}{c|}{} & \multicolumn{11}{l}{{look, be, take, deal, work, make, consider, do, care, \textbf{try}}} & & \multicolumn{11}{l}{{not, move, be, avoid, never, hardly, no, do, \textbf{refrain}, \textbf{fail}}}\\
\multicolumn{1}{c|}{} & \multicolumn{11}{l}{{look, be, take, deal, make, work, \textbf{see}, do, get, consider}} & & \multicolumn{11}{l}{{not, move, be, avoid, never, no, hardly, do, make, \textbf{go}}} \\ [0.5ex]
    \bottomrule 
  \end{tabular} 
  \label{tab:examples}
\end{table*}

To further analyze the capability in exploration, we attempt to count the tokens in the top-$k$ which have never been included in previous epochs. 
    \begin{align}
        \Big|\{\yb_t \in \text{Top-}k \,\, & \pb_{
        \theta}^i (\yb_t | \hb_{t}) \} \nonumber \\
        & \& \{\yb_t \notin \text{Top-}k \,\, \pb_{
        \theta}^j (\yb_t | \hb_{t}), \forall j < i \} \Big|  \nonumber
    \end{align}
    
where the superscript $i$ denotes epoch. We randomly sample 2000 sentence pairs from WMT'17 De-En training data. In Fig.\ref{fig:novel}, we see that there are much more novel tokens in KA than that in KD. Table.\ref{tab:examples} demonstrates two examples. a) Given the prefix tokens ``I will'' and the source sentence, KA probably imagines
\begin{align}
    \texttt{try} \text{ to attend?} \quad \texttt{give} \text{ attention?} \quad \texttt{see} \text{ to?} \nonumber
\end{align}
which provide different ways to express the meaning of \texttt{attend}, while KD just try to explore \texttt{concern}. b) when predicting the token \texttt{no}, KA proposes a set of negative tokens
\begin{align}
    \texttt{fail,} \, \texttt{resist,} \, \texttt{refrain} \nonumber
\end{align}
, and even \texttt{go} which may relate to the next token \texttt{move}.

Fig.\ref{fig:ent} shows that the entropy of KA goes up while the entropy of KD dropping during fine-tuning. This also indicates that KA motivates exploration more than KD.

\section{Conclusion}
In this paper we took a close look at how knowledge transfer can be used to improve the capabilities of a neural network model for the sequence generation task (the learner model) using another model which is known to be stronger (the teacher model). While we focused on improving a single learning model from a single (fixed) teacher model, in future work it is worth exploring a joint learning system where all agents are learners but with different roles, where they have to cooperate or compete to accomplish a task.

We explored the details of the learning process when optimizing KL-divergence in forward and backward orders. We found that \texttt{Backward} allows learners to acquire knowledge in a more efficient way, especially in solving sequential decision making problems. Our analysis is general and applicable to other tasks. We believe it would guide us to utilize KL-divergence effectively.

\nocite{langley00}

\bibliography{example_paper}
\bibliographystyle{icml2020}

\appendix
\section{Implicit knowledge transfer}
\subsection{Variational Inference}
\label{sec:vi}
The motivation of VI is to find a simple distribution to approximate the real posterior distribution which is complex and computationally intractable.
\begin{align}
    \kl \big(\underbrace{q(\zb \mid \yb)}_{\text{learner}} \| \underbrace{p(\zb \mid \yb)}_{\text{real posterior as teacher}} \big) \nonumber
\end{align}
where,
\begin{align}
p(\zb \mid \yb) = \frac{p(\yb \mid \zb) p(\zb)}{\sum_z p(\yb \mid \zb) p(\zb)}\quad [\text{Bayes' rule}] \nonumber
\end{align}
Note that, we swap the notation in order to be consistent with the literature.
\subsection{Actor-Critic}
\label{sec:ac}
The Actor-critic algorithm with maximum entropy can be written as maximize
\begin{align}
    & \mathbb{E}_{p(\ab \mid \sb)} \big[ Q(s, a) \big] + \mathbb{H} \big[p(\ab \mid \sb) \big] \nonumber
\end{align}
which is equivalent to
\begin{align}
& \mathbb{E}_{p(\ab \mid \sb)} \Big[ \log \frac{\exp \big(Q(s, a) \big)}{\sum_a \exp \big(Q(s, a) \big)} \Big] + \mathbb{H} \big[p(\ab \mid \sb) \big] \nonumber \\
= & \mathbb{E}_{p(\ab \mid \sb)} \Big[ \log q(\ab \mid \sb) \Big] + \mathbb{H} \big[p(\ab \mid \sb) \big] \nonumber\\
= & - \kl \big(p(\ab \mid \sb) \| \underbrace{q(\ab \mid \sb)}_{\text{Induced by Q function}} \big) \nonumber
\end{align}

\section{Learning strategy}
\subsection{Derivatives}
\label{sec:derivative}
\begin{itemize}
    \item[(a)] \texttt{Forward.} The goal is to minimize
\begin{align}
\sum \qb(x) \log (\frac{\qb(x)}{\pb(x)}), \,\, s.t., \sum \pb(x)=1 \label{eqn:qp}
\end{align}
We write the Lagrangian for Eqn.\ref{eqn:qp} as
\begin{align}
L_{\qb \| \pb}(x, \lambda) = \sum \qb(x) \log (\frac{\qb(x)}{\pb(x)}) + \lambda \big( \sum \pb(x)-1 \big) \nonumber
\end{align}
Method of Lagrangian multipliers involves setting the
derivative of $L_{\qb \| \pb}$ w.r.t $\pb(x)$ to $0$,
\begin{align}
   G_{\qb \| \pb} = \frac{\partial L_{\qb \| \pb}}{\partial \pb(x)} = \lambda - \frac{\qb(x)}{\pb(x)} = 0 \nonumber
\end{align}
Using the fact that $\sum \pb(x)=1$, we can show that $\lambda=1$.

\item[(b)] \texttt{Backward.} The goal is to minimize
\begin{align}
\sum \pb(x) \log (\frac{\pb(x)}{\qb(x)}), \,\, s.t., \sum \pb(x)=1 \label{eqn:pq}
\end{align}
We write the Lagrangian for Eqn.\ref{eqn:pq} as
\begin{align}
L_{\pb \| \qb}(x, \lambda) = \sum \pb(x) \log (\frac{\pb(x)}{\qb(x)}) + \lambda \big( \sum \pb(x)-1 \big) \nonumber
\end{align}
Method of Lagrangian multipliers involves setting the
derivative of $L_{\pb \| \qb}$ w.r.t $\pb(x)$ to $0$,
\begin{align}
    G_{\pb \| \qb} = \frac{\partial L_{\pb \| \qb}}{\partial \pb(x)} = 1+\lambda + \log \frac{\pb(x)}{\qb(x)} = 0 \nonumber
\end{align}
Using the fact that $\sum \pb(x)=1$, we can show that $\lambda=-1$.
\end{itemize}

\subsection{Property}
\label{sec:property}
We'll prove the key properties in Sec.\ref{sec:dev}.
\begin{itemize}
    \item[(a)] When $\qb(x) > \pb(x)$, we have $1-\frac{\qb(x)}{\pb(x)} < 0$ and $\log \frac{\pb(x)}{\qb(x)} < 0$. And, when $\qb(x) < \pb(x)$, we have $1-\frac{\qb(x)}{\pb(x)} > 0$ and $\log \frac{\pb(x)}{\qb(x)} > 0$.
    \item[(b)] Let's first consider the function $z - \log z$
        \begin{align}
            z - \log z  \begin{cases}
          >1  & \quad \text{if } z \neq 1\\
          =1  & \quad \text{if } z = 1 
         \end{cases}
         \label{eqn:z}
        \end{align}
        It's easy to prove because when $z<1$, the gradient $1-\frac{1}{z}<0$ and when $z>1$, the gradient $1-\frac{1}{z}>0$. Thus, the function reaches the minimum value 1 at $z=1$.
   \begin{itemize}
       \item[(I)] When $\qb(x) > \pb(x)$, we have
    \begin{align}
        \big|G_{\qb \| \pb}(x) \big| - \big| G_{\pb \| \qb}(x) \big| &= \big| 1- \frac{\qb(x)}{\pb(x)}\big| - \big| \log \frac{\pb(x)}{\qb(x)} \big| \\
    &= \frac{\qb(x)}{\pb(x)} - \log \frac{\qb(x)}{\pb(x)} -1 > 0 \nonumber
    \end{align}
    \item[(II)] When $\qb(x) < \pb(x)$, we have
    \begin{align}
    \big|G_{\qb \| \pb}(x) \big| - \big| G_{\pb \| \qb}(x) \big| &= \big| 1- \frac{\qb(x)}{\pb(x)}\big| - \big| \log \frac{\pb(x)}{\qb(x)} \big| \\
    & = 1- (\frac{\qb(x)}{\pb(x)} - \log \frac{\qb(x)}{\pb(x)}) < 0 \nonumber
    \end{align}
   \end{itemize}
\end{itemize}


\end{document}

%% file: definitions.tex
\newcommand{\xb}{\mathbf{x}}

\newcommand{\yb}{\mathbf{y}}
\newcommand{\zb}{\mathbf{z}}
\newcommand{\ab}{\mathbf{a}}

\newcommand{\hb}{\mathbf{h}}

\renewcommand{\sb}{\mathbf{s}}

